\documentclass[a4paper,twoside]{article}

\usepackage{epsfig}
\usepackage{subcaption}
\usepackage{calc}
\usepackage{amssymb}
\usepackage{amstext}
\usepackage{amsmath}
\usepackage{amsthm}
\usepackage{multicol}
\usepackage{pslatex}
\usepackage{apalike}
\usepackage{algorithm2e}
\usepackage[bottom]{footmisc}

\usepackage{amsmath,amsfonts}
\usepackage{array}
\usepackage{textcomp}
\usepackage{stfloats}
\usepackage[hyphens]{url}
\usepackage{verbatim}
\usepackage{graphicx}
\usepackage{cite}

\usepackage{amsmath,amssymb,amsfonts}
\usepackage{textcomp}
\usepackage{xcolor}
\usepackage{comment}
\usepackage{multirow}
\usepackage{siunitx,booktabs}
\usepackage[export]{adjustbox}
\usepackage{makecell}
\usepackage{footmisc}
\usepackage{xurl}

\usepackage{SCITEPRESS}     

\begin{document}

\title{Caption-Matching: A Multimodal Approach for Cross-Domain Image Retrieval}

\author{\authorname{Lucas Iijima\sup{1}, Nikolaos Giakoumoglou\sup{1} and Tania Stathaki\sup{1}}
\affiliation{\sup{1}Department of Electrical and Electronic Engineering, Imperial College London, London, UK}
\email{lk422@ic.ac.uk, \{n.giakoumoglou23, t.stathaki\}@imperial.ac.uk}
}


\abstract{Cross-Domain Image Retrieval (CDIR) is a challenging task in computer vision, aiming to match images across different visual domains such as sketches, paintings, and photographs. Existing CDIR methods rely either on supervised learning with labeled cross-domain correspondences or on methods that require training or fine-tuning on target datasets, often struggling with substantial domain gaps and limited generalization to unseen domains. This paper introduces a novel CDIR approach that incorporates textual context by leveraging publicly available pre-trained vision-language models. Our method, Caption-Matching (CM), uses generated image captions as a domain-agnostic intermediate representation, enabling effective cross-domain similarity computation without the need for labeled data or further training. We evaluate our method on standard CDIR benchmark datasets, demonstrating state-of-the-art performance in plug-and-play settings with consistent improvements on Office-Home and DomainNet over previous methods. We also demonstrate our method's effectiveness on a dataset of AI-generated images from Midjourney, showcasing its ability to handle complex, multi-domain queries.}

\onecolumn \maketitle \normalsize \setcounter{footnote}{0} \vfill


\section{\uppercase{Introduction}}
\label{sec:intro}

Image Retrieval (IR) is a fundamental computer vision task that involves finding images similar to a provided query from a large database. The applications of IR range from content-based search engines to visual question answering systems \cite{antol2015vqa}. As the field has progressed, the challenge has expanded beyond single-domain retrieval to Cross-Domain Image Retrieval (CDIR), where the query and target images may belong to different visual domains such as sketches, paintings, or photographs.

Cross-domain image retrieval (CDIR) poses significant challenges due to the large domain gap between query and target images. Traditional image retrieval methods often fail in cross-domain settings because discriminative features within a single domain may not generalize across domains \cite{wang2017adversarial}. This domain gap, manifested as substantial differences in visual characteristics between domains, frequently results in poor retrieval performance.

Prior CDIR approaches have largely relied on supervised learning, requiring extensive labeled data and cross-domain correspondences \cite{sangkloy2016sketchy, yu2016sketch, song2017deep}, but collecting such annotations is labor-intensive and limits scalability. Domain adaptation techniques attempt to align feature spaces across domains but often struggle when the visual domains are highly disparate \cite{tzeng2017adversarial}. Deep learning approaches that learn domain-invariant features can improve alignment, yet they typically demand significant computational resources and may fail to generalize to unseen domains \cite{ganin2016domain}. Unsupervised and self-supervised methods have been proposed to mitigate the reliance on labeled data \cite{kim2021cds, li2021prototypical, hu2022feature, wang2023correspondence}, but generally require training or fine-tuning on the target datasets and may not fully overcome the domain gap.

Recent advancements in vision-language models, such as CLIP \cite{radford2021learning}, ALIGN \cite{jia2021scaling}, and BLIP-2 \cite{li2023blip}, have opened new possibilities to tackle CDIR tasks. These models, pre-trained on large datasets of image-text pairs, have demonstrated remarkable zero-shot generalization across various vision tasks. However, their potential for CDIR remains largely unexplored.

In this paper, we introduce a novel approach to CDIR that leverages the capabilities of Large Language Models (LLMs) and Vision Transformers (ViTs) without the need for domain-specific annotations or fine-tuning. Our method, termed Caption-Matching (CM), introduces caption-based semantic alignment, a conceptual shift that treats text as a domain-agnostic representation for cross-domain retrieval. Unlike prior methods that operate solely in the visual embedding space by learning domain-invariant features or aligning feature distributions, CM employs natural language descriptions to bridge domain gaps in a fundamentally different manner. By mapping images from various domains to a shared semantic space defined by captions, CM not only facilitates effective cross-domain retrieval without explicit domain alignment but also captures high-level semantic similarities across domains. This approach leverages the combined semantic understanding of LLMs and visual comprehension of ViTs, allowing our system to handle multiple domains simultaneously without requiring separate models or extensive fine-tuning for each domain pair, setting it apart from conventional techniques. The key contributions of our work are as follows:

\begin{itemize}
    \item We introduce a novel CDIR framework that operates entirely without additional training, leveraging pre-trained vision-language models to eliminate the need for labeled cross-domain data or correspondence supervision. Our method uses generated image captions as a domain-agnostic intermediate representation, enabling effective cross-domain similarity computation.
    \item We demonstrate the effectiveness of our approach on DomainNet and Office-Home datasets, surpassing state-of-the-art performance in CDIR tasks.We further validate our method on a diverse dataset of Midjourney-generated images, showing its robustness in complex, multi-domain retrieval scenarios. 
\end{itemize}


\section{\uppercase{Related Work}} \label{sec:related}

\subsection{Image Retrieval}

Image retrieval (IR) is a fundamental task in computer vision, typically involving the retrieval of images given an image query (\textit{image-image} retrieval) \cite{noh2017large}. Traditional approaches often follow a coarse-to-fine strategy, combining global retrieval with local feature aggregation and spatial verification. Deep learning and CNN-based methods have largely replaced hand-crafted features, yielding significant performance improvements \cite{noh2017large,radenovic2018fine}. Notable methods include DELF and DELG \cite{noh2017large}, which learn local and global features jointly, GeM pooling \cite{radenovic2018fine} for attentive aggregation, SOLAR \cite{ng2020solar} for self-attentive feature aggregation, and DOLG \cite{yang2021dolg}, which fuses local and global cues. Despite these advances, most IR systems remain sensitive to visually similar distractors, as global embeddings often rely on geometric consistency during re-ranking \cite{lee2023revisiting}.

Retrieval paradigms have progressively evolved to embed natural language as a core interface for query formulation. Text-based image retrieval (TBIR) \cite{vendrow2024inquire} uses textual queries to find images (\textit{text-image} retrieval), while other methods \cite{song2017fine, sangkloy2022sketch, jia2021scaling} combine an image with its description as a query (\textit{[image+text]-image} retrieval) to improve similarity comparisons. In contrast, our method introduces a novel paradigm in which an image query retrieves semantically associated captions, each linked to an image, effectively performing image retrieval while leveraging a domain-agnostic, text-based intermediate (\textit{image-text} retrieval). Importantly, it operates without explicit domain alignment or labeled cross-domain pairs and can handle multiple domains simultaneously, unlike traditional methods that require separate models or fine-tuning for each domain combination.

\subsection{Cross-Domain Image Retrieval}

Cross-domain image retrieval (CDIR) extends the challenge of image retrieval by searching across diverse domains, such as sketches, cartoons, paintings, and photographs. The primary challenge in CDIR is bridging the domain gap between query and database images. Early approaches leveraged category information for discriminative feature extraction or minimized losses like triplet \cite{yu2016sketch} and HOLEF \cite{song2017deep} for cross-domain pairing. However, these methods often require labor-intensive cross-domain correspondence annotations, limiting their practical applications \cite{sangkloy2016sketchy}.

Recent efforts have explored unsupervised or self-supervised strategies to reduce reliance on labeled data. CDS \cite{kim2021cds} combines in-domain instance discrimination with cross-domain matching to learn domain-invariant representations. ProtoNCE \cite{li2021prototypical} introduces prototypes as latent semantic anchors within a prototypical contrastive framework. PCS \cite{yue2021prototypical} extends prototypical learning to few-shot domain adaptation through instance–prototype alignment. CCL \cite{hu2022feature} uses cluster-wise contrastive learning with a distance-to-distance objective for semantic alignment. CoDA \cite{wang2023correspondence} projects images into a shared subspace via correspondence-free alignment guided by self-matching and classifier-level consistency. While diverse in approach, all of these methods require training or fine-tuning on the target CDIR datasets, even when initialized with large pretrained backbones. In contrast, our caption-matching framework advances CDIR by operating entirely without training or fine-tuning on the target datasets.

\subsection{Vision-Language Foundation Models}

Vision-language models, such as CLIP \cite{radford2021learning} and ALIGN \cite{jia2021scaling}, have emerged as powerful foundation models by pre-training image and language encoder pairs on large-scale image-caption datasets. These models have demonstrated remarkable zero-shot generalization capabilities across various tasks, including image retrieval, classification, and visual question answering \cite{zhou2022learning,song2022clip}.

Following CLIP's success, numerous vision-language foundation models have been developed, incorporating larger datasets, novel architectures, and advanced training objectives \cite{alayrac2022flamingo,li2022blip,singh2022flava}. BLIP-2 \cite{li2023blip}, which introduces a querying transformer to mediate between frozen pre-trained image encoders and LLMs, achieving state-of-the-art performance on various vision-language tasks. Although CLIP and BLIP-2 have been widely used for retrieval and captioning, their joint use as an intermediate semantic space for cross-domain retrieval remains unexplored.

In this work, we present a novel approach to adapt vision-language foundation models for cross-domain image retrieval tasks without requiring a dedicated CDIR dataset. Our caption-matching method uniquely combines the strengths of LLMs and ViTs, allowing it to capture high-level semantic similarities across diverse visual domains. 


\section{\uppercase{Methodology}} \label{sec:method}

\begin{figure*}[htbp]
    \centerline{\includegraphics[width=0.93\textwidth]{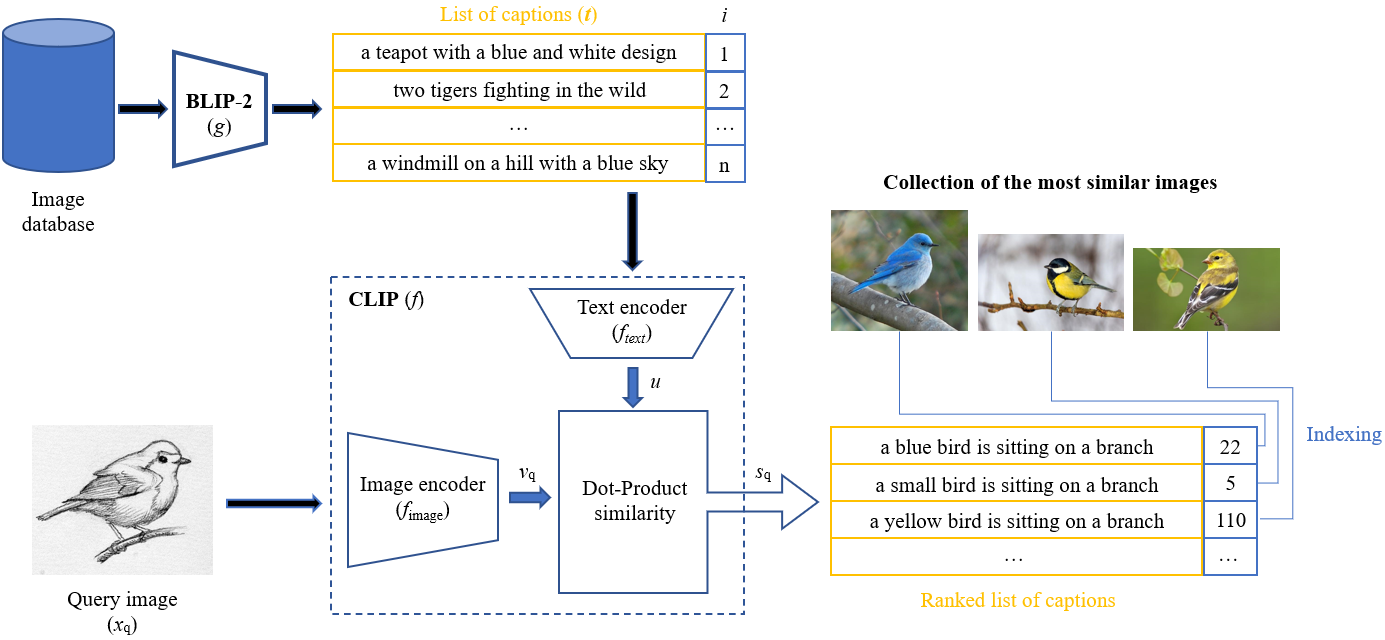}}
    \caption{Overview of the proposed Caption-Matching (CM) method. The process begins with the BLIP-2 model \( g \) generating captions for each image in the database, where \( g(x_i) = t_i \) produces textual captions \( t_i \) from images \( x_i \). These captions are then processed through CLIP's text encoder to obtain text embeddings \( u_i = f_{\text{text}}(t_i) \). Simultaneously, the query image \( x_q \) is encoded by CLIP's image encoder \( f_{\text{image}} \) to produce a visual embedding \( v_q = f_{\text{image}}(x_q) \). A dot-product similarity score \( S_{iq} = v_q \cdot u_i \) is calculated between \( v_q \) and each \( u_i \), resulting in a list of similarity scores \(s_q \). The relevance of each caption to the query image is ranked by sorting \( s_q \) and the corresponding images can be retrieved based on the highest scores.}
    \label{fig:fig1}
\end{figure*}

Here, we present our method, Caption-Matching (CM), which utilizes pre-trained vision-language models to facilitate cross-domain image retrieval through a domain-agnostic, caption-based semantic embedding. This approach allows CM to effectively bridge the visual domain gap, enhancing retrieval accuracy without the need for extensive labeled datasets or domain-specific tuning. Figure \ref{fig:fig1} depicts an overview of the proposed approach.

\subsection{Problem Statement}

Consider a dataset distributed across a set of domains \( S \), where \( A \subset S \) and \( B \subset S \) represent specific subsets associated with the query and target domains, respectively. The task of cross-domain image retrieval involves querying each image from subset \( D_A \) in domain \( A \) and retrieving a selection of the top-\( k \) similar images from subset \( D_B \) in domain \( B \). This operation is formally denoted as \( A \rightarrow B \). Evaluation is conducted at the category level, where retrieved images are deemed correct if they match the category of the query image. The challenge lies in effectively bridging the visual and semantic gaps between domains \( A \) and \( B \) without relying on labeled correspondences, requiring robust feature extraction that can generalize across these varied domains.

\subsection{The Caption-Matching Method}

Aiming to implement a domain-agnostic retrieval system, we introduce the Caption-Matching (CM) method, which leverages the advanced capabilities of LLMs and ViTs to analyze and synchronize data extracted from both image and text modalities. This method employs two principal models: \( f \), a CLIP-based model that comprises \( f_{\text{image}} \) for image encoding and \( f_{\text{text}} \) for text encoding, and \( g \), a BLIP-2-based model for generating textual captions from images.

Initially, each image \( x_i \) from the target domain \( B \) is processed by \( g \) to generate a descriptive caption \( t_i \). These images are transformed into textual representations \( t_i = g(x_i) \), converting visual data into a semantically enriched text format. Subsequently, in the comparison phase, each image \( x_j \) from the query domain \( A \) is encoded into its visual embedding \( v_j = f_{\text{image}}(x_j) \) by the CLIP model. Simultaneously, the text encoder \( f_{\text{text}} \) is used to encode textual captions into embeddings. The matching process then computes the dot-product similarity scores \( S_{ij} = v_j \cdot u_i \) between \( v_j \) and each \( u_i \), where \( u_i \) is derived from \( f_{\text{text}}(t_i) \), following CLIP's standard formulation. This metric assesses the relevance between the query images and the captions of the target images, facilitating a ranked retrieval of the most semantically aligned images. The final output is a sorted list of images from domain \( B \) that best correspond to the query image in domain \( A \), thus bridging the semantic gap across domains via an effective integration of visual and linguistic analysis. 

Figure \ref{fig:fig1} shows the diagram for the method proposed, where a BLIP-2 model ($g$) acts as an image captioning model and a CLIP model ($f$) acts as an image-text matching model. The combined implementation of an image captioning and an image-text matching model allows image representation using natural language, facilitating the seamless embedding of contextual information.

\begin{figure}[htbp]
    \centerline{\fbox{\includegraphics[width=0.38\textwidth]{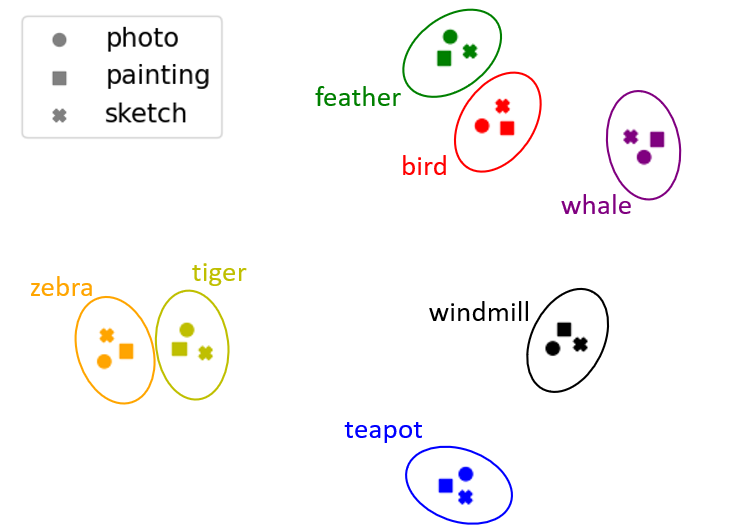}}}
    \caption{t-SNE visualization of CLIP text embeddings illustrating clustering capability across domains. In this experiment, captions were artificially constructed using the template "\{Domain\} of a \{Class\}" (e.g., “painting of a bird”), showing how semantic similarities are preserved while domain-specific identifiers are abstracted away, which facilitates domain-agnostic categorization.}
    \label{fig:fig2}
\end{figure}

In the context of CDIR, a notable feature of CLIP when pre-trained on a sufficiently large and diverse dataset is its ability to bridge the domain gap without requiring modifications to the embedding space. Figure \ref{fig:fig2} illustrates this with a t-SNE \cite{maaten2008tsne} projection of text embeddings from selected examples. It shows that CLIP’s text encoder clusters captions based on their semantic context while disregarding domain-specific terms such as “photo”, “painting”, and “sketch.” This enables the corresponding image embeddings to be accurately matched with the correct image category across different domains. Hence, we choose captions as intermediates because text abstracts away stylistic domain characteristics while preserving semantic identity, making it suitable for CDIR.

\section{\uppercase{Experiments}} \label{sec:exp}

\subsection{Implementation Details} 

\textbf{Models.} We use two publicly available pre-trained models: BLIP-2\footnote{https://huggingface.co/Salesforce/blip2-opt-2.7b} for image captioning and CLIP\footnote{https://huggingface.co/laion/CLIP-ViT-H-14-laion2B-s32B-b79K} for image-text matching. BLIP-2 combines a frozen ViT image encoder with a pre-trained OPT language model and was trained on a mixture of human-annotated datasets (COCO, Visual Genome) and large-scale web datasets (CC3M, CC12M, SBU, LAION-400M). CLIP uses a ViT-Huge image encoder with a GPT-2 text encoder and was trained on the LAION-2B-en dataset. Both models are used with default inference parameters and their publicly available pre-trained weights without modification.

\textbf{Datasets.} We evaluate our method using two diverse datasets. (1) \textbf{Office-Home} \cite{venkateswara2017deep} comprises approximately 15,500 manually annotated images sourced from the web, featuring four distinct domains: \textit{art} (\textbf{Ar}), \textit{clipart} (\textbf{Cl}), \textit{product} (\textbf{Pr}), and \textit{real world} (\textbf{Rw})—spanning across 65 categories of common everyday objects such as chairs, keyboards, and bikes. (2) \textbf{DomainNet} \cite{peng2019moment} contains around 600,000 manually annotated images collected from the web, including six diverse domains: \textit{clipart} (\textbf{C}), \textit{infograph} (\textbf{I}), \textit{painting} (\textbf{P}), \textit{quickdraw} (\textbf{Q}), \textit{real} (\textbf{R}), and \textit{sketch} (\textbf{S}), spanning across 345 categories ranging from tools and clothing to natural objects and human body categories.

\textbf{Baselines.} We compare our method to state-of-the-art CDIR approaches: CDS \cite{kim2021cds}, ProtoNCE \cite{li2021prototypical}, PCS \cite{yue2021prototypical}, CCL \cite{hu2022feature}, and CoDA \cite{wang2023correspondence}. These methods require training on cross-domain target datasets, whereas CM requires no task-specific training or fine-tuning, yielding a direct comparison between training-free and training-based approaches. Furthermore, while all methods require feature extraction backbones, our goal is not to compare raw model capacity, but rather to evaluate whether caption-based representations provide an alternative paradigm for cross-domain retrieval. Accordingly, our results highlight methodological differences rather than variations in model scale.

\textbf{Evaluation.} We follow the experimental setup and metrics of \cite{hu2022feature, wang2023correspondence}. On Office-Home, we evaluate all query-target domain pairs using precision scores P@1, P@5, P@15, and mAP@All. On DomainNet, we consider categories with over 200 samples and evaluate all domain pairs using P@50, P@100, and P@200.


\subsection{Results} 

\begin{figure}[htbp]
    \scriptsize
    \centering 
    \begin{tabular}{@{} c | c@{\hskip 0.0in} c@{}} 
    Query & Retrieved Results & Method \\
    \hline \\[-5.5pt]
    \multirow[c]{3}{*}{\includegraphics[height=.25in, width=.25in, margin=0pt 0ex 0pt -10ex]{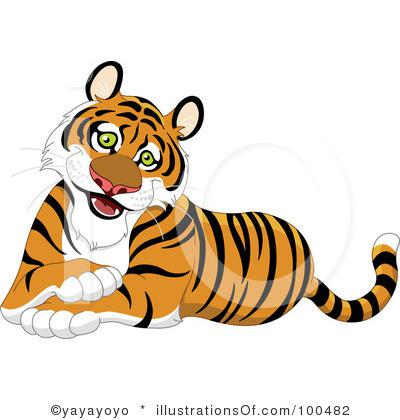}} & 
    \includegraphics[width=0.355\textwidth, margin=0pt 0ex 0pt 0ex]{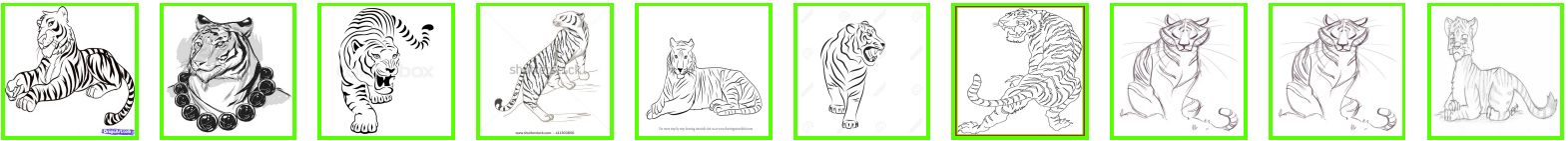} & \makecell{\\[-6ex] CCL} \\
    & \includegraphics[width=0.355\textwidth, margin=0pt 0ex 0pt 0ex]{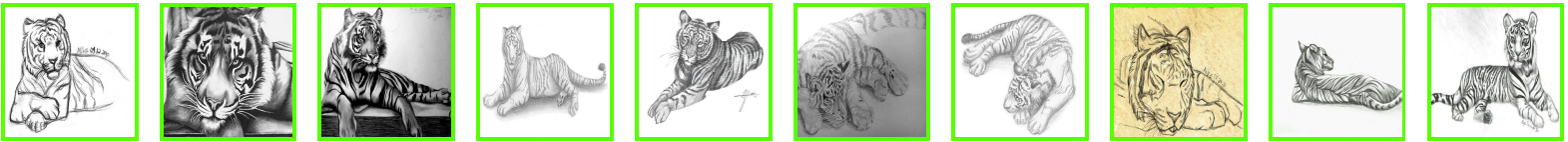} & \makecell{\\[-6ex] CM} \\ 
    \hline \\[-5.5pt]
    
    \multirow[c]{3}{*}{\includegraphics[height=.25in, width=.25in, margin=0pt 0ex 0pt -10ex]{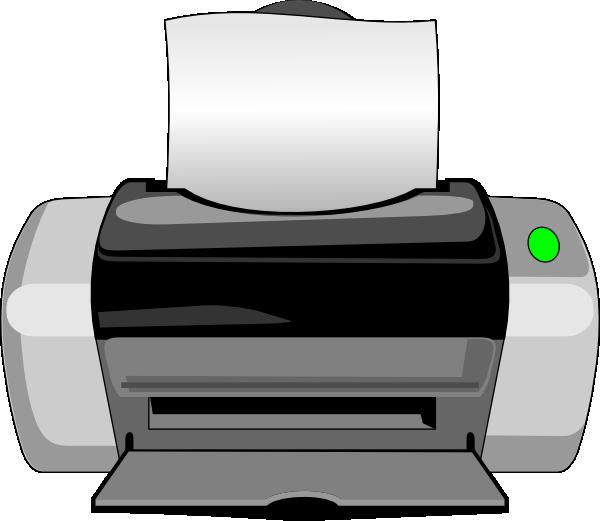}} & 
    \includegraphics[width=0.355\textwidth, margin=0pt 0ex 0pt 0ex]{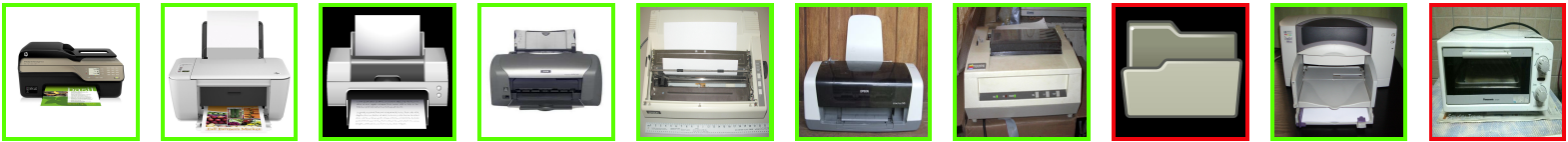} & \makecell{\\[-6ex] CCL} \\
    & \includegraphics[width=0.355\textwidth, margin=0pt 0ex 0pt 0ex]{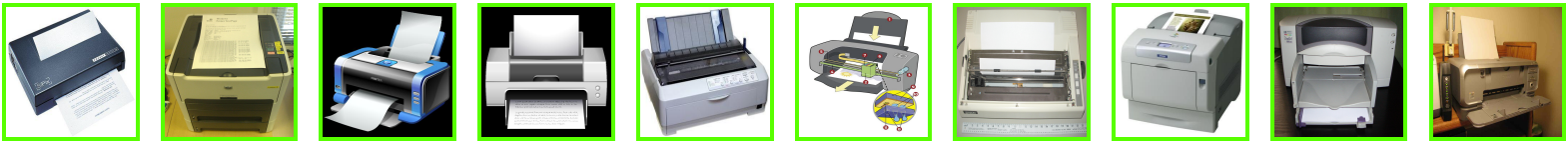} & \makecell{\\[-6ex] CM} \\ 
    \hline \\[-5.5pt]
    
    \multirow[c]{3}{*}{\includegraphics[height=.25in, width=.25in, margin=0pt 0ex 0pt -10ex]{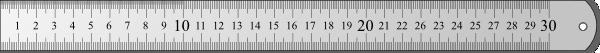}} & 
    \includegraphics[width=0.355\textwidth, margin=0pt 0ex 0pt 0ex]{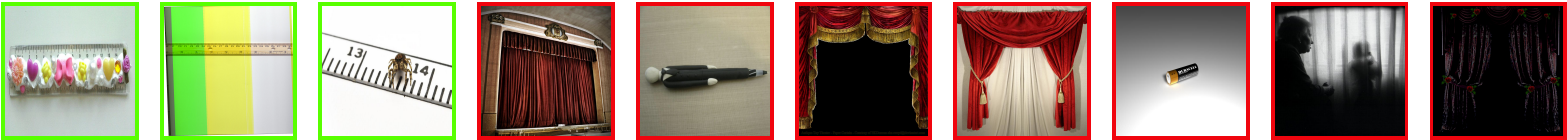} & \makecell{\\[-6ex] CoDA} \\
    & \includegraphics[width=0.355\textwidth, margin=0pt 0ex 0pt 0ex]{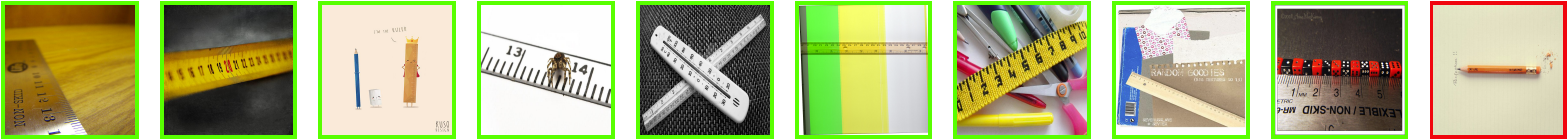} & \makecell{\\[-6ex] CM} \\ 
    \hline
\end{tabular} 

\caption{Qualitative comparison of the top-10 retrieved results. From top to bottom, the domain pair are C-S (DomainNet), Cl-Rw (Office-Home) and Cl-Ar (Office-Home). Correctly retrieved images are outlined in green, while incorrectly retrieved images are outlined in red.}
\label{fig:fig3}
\end{figure}

Table \ref{tab:tab1} summarizes the experimental results and comparisons with state-of-the-art methods. Our approach, which combines BLIP-2 for image captioning and CLIP for image-text matching, substantially outperforms existing techniques. On DomainNet and Office-Home, it nearly doubles the precision of prior methods across P@50, P@100, P@200 (DomainNet) and P@1, P@5, P@15 (Office-Home). It also consistently achieves higher mAP scores across all Office-Home domain pairs. These large gains arise because captions often encode semantic cues that remain invariant across visual domains, enabling CLIP’s text encoder to perform more stable matching than visual features alone. Complementing the quantitative analysis, Figure \ref{fig:fig3} presents a qualitative comparison, illustrating the superiority of CM and revealing insights not captured by standard precision metrics.

\begin{table*}[t]
    \centering
    \caption{Comparison of retrieval performance between our method (CM) and state-of-the-art methods on the DomainNet and Office-Home datasets. Domain pairs are denoted by abbreviations (e.g., C-S for Clipart to Sketch). Bold values indicate the best performance for each metric and domain pair.}
    \label{tab:tab1}
    \scriptsize
    \setlength{\tabcolsep}{4pt}
    \renewcommand{\arraystretch}{0.8}
    \begin{tabular}{ccccccccccccccc}
        \hline
        \multicolumn{14}{c}{\textbf{DomainNet}} \\
        \textbf{} & \textbf{C-S} & \textbf{S-C} & \textbf{I-R} & \textbf{R-I} & \textbf{I-S} & \textbf{S-I} & \textbf{P-C} & \textbf{C-P} & \textbf{P-Q} & \textbf{Q-P} & \textbf{Q-R} & \textbf{R-Q} & \textbf{Avg} \\
        \hline
        \multicolumn{14}{c}{P@50} \\
        \hline
        CDS \cite{kim2021cds} & 0.458 & 0.591 & 0.285 & 0.567 & 0.306 & 0.463 & 0.632 & 0.378 & 0.188 & 0.214 & 0.193 & 0.154 & 0.369 \\
        ProtoNCE \cite{li2021prototypical} & 0.468 & 0.545 & 0.284 & 0.570 & 0.282 & 0.398 & 0.554 & 0.391 & 0.216 & 0.240 & 0.264 & 0.251 & 0.372 \\
        PCS \cite{yue2021prototypical} & 0.510 & 0.597 & 0.306 & 0.554 & 0.303 & 0.426 & 0.635 & 0.488 & 0.251 & 0.240 & 0.348 & 0.290 & 0.412 \\
        CCL \cite{hu2022feature} & 0.563 & 0.631 & 0.355 & 0.577 & 0.313 & 0.437 & 0.664 & 0.526 & 0.397 & 0.334 & 0.428 & 0.419 & 0.470 \\
        CM (ours) & \textbf{0.968} & \textbf{0.939} & \textbf{0.693} & \textbf{0.884} & \textbf{0.722} & \textbf{0.879} & \textbf{0.971} & \textbf{0.931} & \textbf{0.832} & \textbf{0.546} & \textbf{0.522} & \textbf{0.792} & \textbf{0.807} \\
        \hline
        \multicolumn{14}{c}{P@100} \\
        \hline
        CDS\cite{kim2021cds} & 0.424 & 0.488 & 0.279 & 0.398 & 0.295 & 0.361 & 0.473 & 0.352 & 0.189 & 0.214 & 0.191 & 0.156 & 0.318 \\
        ProtoNCE\cite{li2021prototypical} & 0.427 & 0.450 & 0.285 & 0.418 & 0.268 & 0.320 & 0.437 & 0.359 & 0.212 & 0.228 & 0.257 & 0.248 & 0.326 \\
        PCS \cite{yue2021prototypical} & 0.469 & 0.507 & 0.303 & 0.421 & 0.284 & 0.341 & 0.532 & 0.462 & 0.246 & 0.232 & 0.339 & 0.289 & 0.369 \\
        CCL \cite{hu2022feature} & 0.527 & 0.573 & 0.352 & 0.467 & 0.293 & 0.361 & 0.568 & 0.501 & 0.386 & 0.338 & 0.428 & 0.421 & 0.435 \\
        CM (ours) & \textbf{0.962} & \textbf{0.919} & \textbf{0.694} & \textbf{0.797} & \textbf{0.711} & \textbf{0.796} & \textbf{0.961} & \textbf{0.921} & \textbf{0.812} & \textbf{0.535} & \textbf{0.501} & \textbf{0.793} & \textbf{0.784} \\
        \hline
        \multicolumn{14}{c}{P@200} \\
        \hline
        CDS\cite{kim2021cds} & 0.372 & 0.374 & 0.275 & 0.264 & 0.270 & 0.273 & 0.329 & 0.328 & 0.179 & 0.195 & 0.187 & 0.158 & 0.267 \\
        ProtoNCE\cite{li2021prototypical} & 0.364 & 0.351 & 0.285 & 0.303 & 0.242 & 0.248 & 0.326 & 0.321 & 0.206 & 0.216 & 0.244 & 0.238 & 0.279 \\
        PCS \cite{yue2021prototypical} & 0.402 & 0.394 & 0.297 & 0.308 & 0.254 & 0.259 & 0.417 & 0.421 & 0.238 & 0.221 & 0.317 & 0.282 & 0.317 \\
        CCL \cite{hu2022feature} & 0.474 & 0.482 & 0.344 & 0.355 & 0.265 & 0.281 & 0.467 & 0.461 & 0.376 & 0.343 & 0.427 & 0.416 & 0.391 \\
        CM (ours) & \textbf{0.935} & \textbf{0.869} & \textbf{0.690} & \textbf{0.652} & \textbf{0.684} & \textbf{0.658} & \textbf{0.921} & \textbf{0.899} & \textbf{0.751} & \textbf{0.522} & \textbf{0.503} & \textbf{0.747} & \textbf{0.736} \\
        \hline
        \multicolumn{14}{c}{\textbf{Office-Home}} \\
        \textbf{} & \textbf{Ar-Cl} & \textbf{Ar-Pr} & \textbf{Ar-Rw} & \textbf{Cl-Ar} & \textbf{Cl-Pr} & \textbf{Cl-Rw} & \textbf{Pr-Ar} & \textbf{Pr-Cl} & \textbf{Pr-Rw} & \textbf{Rw-Ar} & \textbf{Rw-Cl} & \textbf{Rw-Pr} & \textbf{Avg} \\
        \hline
        \multicolumn{14}{c}{P@1} \\
        \hline
        CDS\cite{kim2021cds} & 0.256 & 0.328 & 0.451 & 0.224 & 0.272 & 0.325 & 0.358 & 0.377 & 0.540 & 0.447 & 0.389 & 0.494 & 0.372 \\
        ProtoNCE\cite{li2021prototypical} & 0.290 & 0.295 & 0.405 & 0.213 & 0.212 & 0.252 & 0.357 & 0.361 & 0.538 & 0.445 & 0.412 & 0.477 & 0.355 \\
        PCS \cite{yue2021prototypical} & 0.312 & 0.333 & 0.417 & 0.245 & 0.264 & 0.291 & 0.392 & 0.395 & 0.564 & 0.450 & 0.406 & 0.499 & 0.381 \\
        CCL \cite{hu2022feature} & 0.327 & 0.354 & 0.451 & 0.273 & 0.278 & 0.333 & 0.425 & 0.423 & 0.574 & 0.480 & 0.447 & 0.517 & 0.407 \\
        CM (ours) & \textbf{0.783} & \textbf{0.718} & \textbf{0.807} & \textbf{0.678} & \textbf{0.711} & \textbf{0.729} & \textbf{0.843} & \textbf{0.883} & \textbf{0.924} & \textbf{0.845} & \textbf{0.848} & \textbf{0.869} & \textbf{0.803} \\
        \hline
        \multicolumn{14}{c}{P@5} \\
        \hline
        CDS\cite{kim2021cds} & 0.238 & 0.315 & 0.412 & 0.203 & 0.265 & 0.303 & 0.325 & 0.350 & 0.501 & 0.408 & 0.365 & 0.473 & 0.346 \\
        ProtoNCE\cite{li2021prototypical} & 0.262 & 0.279 & 0.364 & 0.174 & 0.206 & 0.227 & 0.306 & 0.340 & 0.482 & 0.393 & 0.377 & 0.448 & 0.321 \\
        PCS \cite{yue2021prototypical} & 0.287 & 0.315 & 0.385 & 0.213 & 0.259 & 0.261 & 0.348 & 0.375 & 0.508 & 0.399 & 0.381 & 0.471 & 0.350 \\
        CCL \cite{hu2022feature} & 0.308 & 0.347 & 0.423 & 0.239 & 0.273 & 0.306 & 0.379 & 0.374 & 0.527 & 0.437 & 0.415 & 0.485 & 0.376 \\
        CM (ours) & \textbf{0.768} & \textbf{0.723} & \textbf{0.791} & \textbf{0.637} & \textbf{0.700} & \textbf{0.712} & \textbf{0.791} & \textbf{0.855} & \textbf{0.878} & \textbf{0.801} & \textbf{0.829} & \textbf{0.855} & \textbf{0.778} \\
        \hline
        \multicolumn{14}{c}{P@15} \\
        \hline
        CDS\cite{kim2021cds} & 0.224 & 0.289 & 0.387 & 0.173 & 0.249 & 0.278 & 0.268 & 0.304 & 0.456 & 0.355 & 0.332 & 0.440 & 0.313 \\
        ProtoNCE\cite{li2021prototypical} & 0.230 & 0.258 & 0.340 & 0.145 & 0.205 & 0.208 & 0.246 & 0.282 & 0.422 & 0.330 & 0.320 & 0.412 & 0.283 \\
        PCS \cite{yue2021prototypical} & 0.261 & 0.295 & 0.362 & 0.175 & 0.249 & 0.240 & 0.288 & 0.328 & 0.454 & 0.340 & 0.341 & 0.437 & 0.314 \\
        CCL \cite{hu2022feature} & 0.287 & 0.326 & 0.401 & 0.205 & 0.260 & 0.281 & 0.314 & 0.337 & 0.479 & 0.384 & 0.374 & 0.450 & 0.341 \\
        CM (ours) & \textbf{0.728} & \textbf{0.694} & \textbf{0.756} & \textbf{0.558} & \textbf{0.656} & \textbf{0.671} & \textbf{0.694} & \textbf{0.804} & \textbf{0.837} & \textbf{0.721} & \textbf{0.788} & \textbf{0.816} & \textbf{0.727} \\
        \hline
        \multicolumn{14}{c}{mAP@All} \\
        \hline
        PCS \cite{yue2021prototypical} & 0.343 & 0.463 & 0.516 & 0.323 & 0.405 & 0.406 & 0.470 & 0.421 & 0.613 & 0.516 & 0.428 & 0.601 & 0.459 \\
        CoDA \cite{wang2023correspondence} & 0.347 & 0.496 & 0.532 & 0.332 & 0.429 & 0.447 & 0.504 & 0.452 & 0.652 & 0.531 & 0.460 & 0.652 & 0.486 \\
        CM (ours) & \textbf{0.528} & \textbf{0.539} & \textbf{0.590} & \textbf{0.452} & \textbf{0.509} & \textbf{0.523} & \textbf{0.571} & \textbf{0.596} & \textbf{0.658} & \textbf{0.596} & \textbf{0.600} & \textbf{0.662} & \textbf{0.569} \\
        \hline
    \end{tabular}
\end{table*}

\subsection{Qualitative Evaluation with Multi-Domain Dataset}

The proposed method is also qualitatively evaluated in an additional experiment with a multi-domain dataset\footnote{Available at \url{https://www.kaggle.com/datasets/iraklip/modjourney-v51-cleaned-data}.} containing AI-generated images. This experiment used only the first 25k images generated with the model’s latest versions, namely Midjourney v5.0, v5.1, and v5.2. Since the images are unlabeled, collections of retrieved images are presented as qualitative results. The aim of this experiment is to show how the CM method performs with complex images from previously unseen categories and ascertain its potential utility as a tool in generative applications.

\begin{figure*}[t]
    \centering 
    \includegraphics[width=0.9\textwidth]{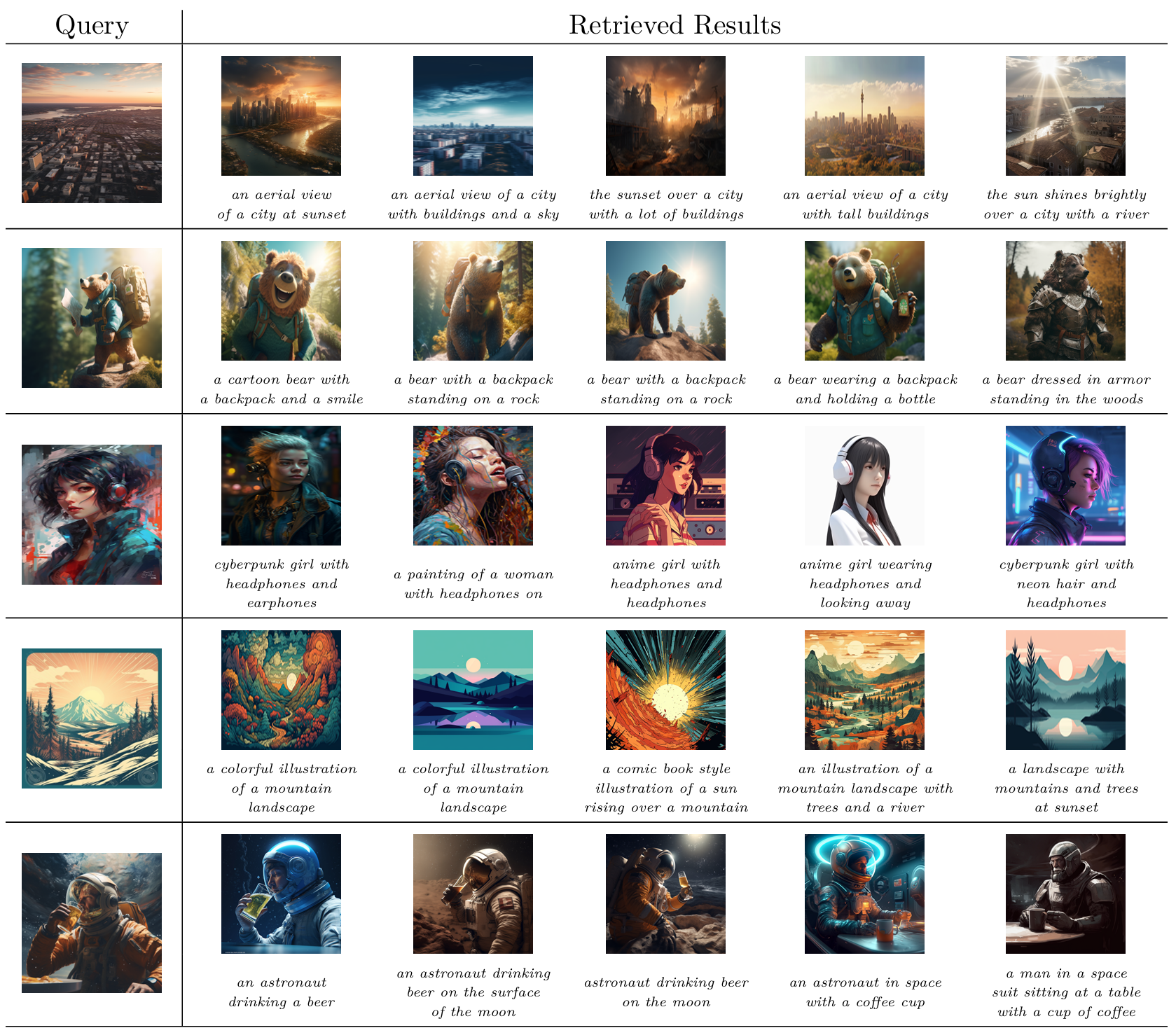}
    \caption{Top-5 retrieval results on Midjourney’s database. Descriptions were generated by BLIP-2 as part of the CM method.}
    \label{fig:fig4}
\end{figure*}

The results presented in Figure \ref{fig:fig4} demonstrate the CM method's robust performance on highly detailed, AI-generated images. Notably, the method exhibits the ability to retrieve images across multiple domains for a single query, for example, pairing realistic images with anime-style illustrations (third query) or drawings with comic-style renderings (fourth query). This capability further emphasizes the domain-agnostic nature of the CM method.

\section{\uppercase{Discussion}} \label{sec:disc}

Our Caption-Matching (CM) approach delivers strong improvements in cross-domain image retrieval while requiring no training or domain-specific adaptation. By leveraging BLIP-2 for caption generation and CLIP for image–text matching, CM inherits rich semantic priors learned from large-scale multimodal datasets. As illustrated in Figure \ref{fig:fig2}, CLIP captures both content and contextual cues, enabling it to bridge domain gaps that challenge traditional visual-only methods.

The use of captions as a domain-agnostic intermediate representation allows CM to encode high-level semantics that remain stable across diverse visual styles. This leads to more meaningful retrieval results, as shown in Figure \ref{fig:fig3}, where CM returns images that align with both the category and contextual attributes of the query. Unlike existing CDIR approaches that rely solely on visual features, CM benefits from the complementary strengths of vision-language modeling.

CM also demonstrates strong scalability, supporting multiple domains without requiring separate models or fine-tuning. Its ViT-based components accept variable input resolutions, avoiding distortions common in CNN-based pipelines. Importantly, in contrast to existing CDIR methods, which are trained directly on the evaluation datasets to learn domain-invariant visual features, CM relies solely on off-the-shelf pre-trained models and achieves superior retrieval performance. This is further reflected in qualitative results such as those in Figure \ref{fig:fig3}, where CM retrieves images that match both the category and pose of the query (e.g., a tiger laying down), capturing contextual semantics that visual-only methods fail to model. These findings highlight the advantages of multimodal representations for CDIR and suggest that further advances in caption generation and large-scale pre-training could yield even greater improvements.


\section{\uppercase{Conclusions}} \label{sec:conc}

We proposed a novel caption-matching approach for cross-domain image retrieval that leverages both language and vision features for the task. The CM method matches a query image with the most suitable descriptions, which are concurrently associated with images from the target database. While most approaches in the literature are limited to operations with visual embeddings, the CM method overcomes the domain gap by strategically integrating CLIP, whose text encoder is able to cluster text descriptions based on image context. It achieves state-of-the-art performance in CDIR and performs remarkably well on AI-generated images, without the need for fine-tuning on specific datasets. Our work highlights the potential of using language as a bridge across visual domains, suggesting new research directions where captions or textual prompts serve as structured intermediates.




\bibliographystyle{apalike}
{\small
\bibliography{references}}

@inproceedings{noh2017large,
  title={Large-scale image retrieval with attentive deep local features},
  author={Noh, Hyeonwoo and Araujo, Andre and Sim, Jack and Weyand, Tobias and Han, Bohyung},
  booktitle={2017 IEEE International Conference on Computer Vision (ICCV)},
  pages={3456-3465},
  year={2017}
}

@article{radenovic2018fine,
  title={Fine-tuning CNN image retrieval with no human annotation},
  author={Radenovi{\'c}, Filip and Tolias, Giorgos and Chum, Ond{\v{r}}ej},
  journal={IEEE Transactions on Pattern Analysis and Machine Intelligence},
  volume={41},
  number={7},
  pages={1655--1668},
  year={2019},
  publisher={IEEE}
}

@inproceedings{ng2020solar,
  title={SOLAR: Second-order loss and attention for image retrieval},
  author={Ng, Tony and Balntas, Vassileios and Tian, Yurun and Mikolajczyk, Krystian},
  booktitle={European Conference on Computer Vision (ECCV)},
  pages={253--270},
  year={2020},
  organization={Springer}
}

@inproceedings{yang2021dolg,
  title={DOLG: Single-stage image retrieval with deep orthogonal fusion of local and global features},
  author={Yang, Min and He, Dongliang and Fan, Miao and Shi, Baorong and Xue, Xuetong and Li, Fu and Ding, Errui and Huang, Jizhou},
  booktitle={2021 IEEE/CVF International Conference on Computer Vision (ICCV)},
  pages={11772--11781},
  year={2021}
}

@inproceedings{lee2023revisiting,
  title={Revisiting Self-Similarity: Structural Embedding for Image Retrieval},
  author={Lee, Seongwon and Lee, Suhyeon and Seong, Hongje and Kim, Euntai},
  booktitle={2023 IEEE/CVF Conference on Computer Vision and Pattern Recognition (CVPR)},
  pages={23412--23421},
  year={2023}
}

@inproceedings{yu2016sketch,
  title={Sketch me that shoe},
  author={Yu, Qian and Liu, Feng and Song, Yi-Zhe and Xiang, Tao and Hospedales, Timothy M and Loy, Chen-Change},
  booktitle={2016 IEEE Conference on Computer Vision and Pattern Recognition (CVPR)},
  pages={799--807},
  year={2016}
}

@inproceedings{song2017deep,
  title={Deep spatial-semantic attention for fine-grained sketch-based image retrieval},
  author={Song, Jifei and Yu, Qian and Song, Yi-Zhe and Xiang, Tao and Hospedales, Timothy M},
  booktitle={2017 IEEE International Conference on Computer Vision (ICCV)},
  pages={5551--5560},
  year={2017}
}

@inproceedings{sangkloy2016sketchy,
  title={The sketchy database: learning to retrieve badly drawn bunnies},
  author={Sangkloy, Patsorn and Burnell, Nathan and Ham, Cusuh and Hays, James},
  booktitle={ACM SIGGRAPH 2016 Papers},
  pages={1--12},
  year={2016}
}

@inproceedings{kim2021cds,
  title={CDS: Cross-domain self-supervised pre-training},
  author={Kim, Donghyun and Saito, Kuniaki and Oh, Tae-Hyun and Plummer, Bryan A and Sclaroff, Stan and Saenko, Kate},
  booktitle={2021 IEEE/CVF International Conference on Computer Vision (ICCV)},
  pages={9123--9132},
  year={2021}
}

@inproceedings{radford2021learning,
  title={Learning transferable visual models from natural language supervision},
  author={Radford, Alec and Kim, Jong Wook and Hallacy, Chris and Ramesh, Aditya and Goh, Gabriel and Agarwal, Sandhini and Sastry, Girish and Askell, Amanda and Mishkin, Pamela and Clark, Jack and Krueger, Gretchen and Sutskever, Ilya},
  booktitle={International Conference on Machine Learning},
  pages={8748--8763},
  year={2021},
  organization={PMLR}
}

@inproceedings{jia2021scaling,
  title={Scaling up visual and vision-language representation learning with noisy text supervision},
  author={Jia, Chao and others},
  booktitle={International Conference on Machine Learning},
  pages={4904--4916},
  year={2021},
  organization={PMLR}
}

@article{zhou2022learning,
  title={Learning to prompt for vision-language models},
  author={Zhou, Kaiyang and Yang, Jingkang and Loy, Chen Change and Liu, Ziwei},
  journal={International Journal of Computer Vision},
  volume={130},
  pages={2337--2348},
  year={2022},
  publisher={Springer}
}

@inproceedings{song2022clip,
  title={CLIP models are few-shot learners: Empirical studies on VQA and visual entailment},
  author={Song, Haoyu and Dong, Li and Zhang, Weinan and Liu, Ting and Wei, Furu},
  booktitle={Proceedings of the 60th Annual Meeting of the Association for Computational Linguistics},
  pages={6088--6100},
  year={2022}
}

@article{alayrac2022flamingo,
  title={Flamingo: a visual language model for few-shot learning},
  author={Alayrac, Jean-Baptiste and others},
  journal={Advances in Neural Information Processing Systems},
  volume={35},
  pages={23716--23736},
  year={2022}
}

@inproceedings{li2022blip,
  title={BLIP: Bootstrapping language-image pre-training for unified vision-language understanding and generation},
  author={Li, Junnan and Li, Dongxu and Xiong, Caiming and Hoi, Steven},
  booktitle={International Conference on Machine Learning},
  pages={12888--12900},
  year={2022},
  organization={PMLR}
}

@inproceedings{singh2022flava,
  title={FLAVA: A foundational language and vision alignment model},
  author={Singh, Amanpreet and Hu, Ronghang and Goswami, Vedanuj and Couairon, Guillaume and Galuba, Wojciech and Rohrbach, Marcus and Kiela, Douwe},
  booktitle={2022 IEEE/CVF Conference on Computer Vision and Pattern Recognition (CVPR)},
  pages={15638--15650},
  year={2022}
}

@article{li2023blip,
  title={BLIP-2: Bootstrapping Language-Image Pre-training with Frozen Image Encoders and Large Language Models},
  author={Li, Junnan and Li, Dongxu and Savarese, Silvio and Hoi, Steven},
  journal={arXiv preprint arXiv:2301.12597},
  year={2023}
}

@article{maaten2008tsne,
  author  = {Laurens van der Maaten and Geoffrey Hinton},
  title   = {Visualizing Data using t-SNE},
  journal = {Journal of Machine Learning Research},
  year    = {2008},
  volume  = {9},
  number  = {86},
  pages   = {2579--2605},
  url     = {http://jmlr.org/papers/v9/vandermaaten08a.html}
}

@inproceedings{antol2015vqa,
  title={Vqa: Visual question answering},
  author={Antol, Stanislaw and Agrawal, Aishwarya and Lu, Jiasen and Mitchell, Margaret and Batra, Dhruv and Lawrence Zitnick, C and Parikh, Devi},
  booktitle={2015 IEEE International Conference on Computer Vision (ICCV)},
  pages={2425--2433},
  year={2015}
}

@inproceedings{wang2017adversarial,
  title={Adversarial cross-modal retrieval},
  author={Wang, Bokun and Yang, Yang and Xu, Xing and Hanjalic, Alan and Shen, Heng Tao},
  booktitle={Proceedings of the 25th ACM International Conference on Multimedia},
  pages={154--162},
  year={2017}
}

@inproceedings{tzeng2017adversarial,
  title={Adversarial discriminative domain adaptation},
  author={Tzeng, Eric and Hoffman, Judy and Saenko, Kate and Darrell, Trevor},
  booktitle={2017 IEEE Conference on Computer Vision and Pattern Recognition (CVPR)},
  pages={2962-2971},
  year={2017}
}

@inproceedings{ganin2016domain,
  title={Domain-adversarial training of neural networks},
  author={Ganin, Yaroslav and Ustinova, Evgeniya and Ajakan, Hana and Germain, Pascal and Larochelle, Hugo and Laviolette, Fran{\c{c}}ois and Marchand, Mario and Lempitsky, Victor},
  booktitle={Domain Adaptation in Computer Vision Applications},
  pages={189--209},
  year={2017},
  organization={Springer}
}

@inproceedings{peng2019moment,
  title={Moment matching for multi-source domain adaptation},
  author={Peng, Xingchao and Bai, Qinxun and Xia, Xide and Huang, Zijun and Saenko, Kate and Wang, Bo},
  booktitle={2019 IEEE/CVF International Conference on Computer Vision (ICCV)},
  pages={1406--1415},
  year={2019}
}

@inproceedings{venkateswara2017deep,
  title={Deep hashing network for unsupervised domain adaptation},
  author={Venkateswara, Hemanth and Eusebio, Jose and Chakraborty, Shayok and Panchanathan, Sethuraman},
  booktitle={2017 IEEE Conference on Computer Vision and Pattern Recognition (CVPR)},
  pages={5018--5027},
  year={2017}
}

@inproceedings{yue2021prototypical,
  title={Prototypical Cross-Domain Self-Supervised Learning for Few-Shot Unsupervised Domain Adaptation},
  author={Yue, Xiangyu and others},
  booktitle={2021 IEEE Conference on Computer Vision and Pattern Recognition (CVPR)},
  year={2021}
}

@inproceedings{song2017fine,
  title = {Fine-Grained Image Retrieval: the Text/Sketch Input Dilemma},
  author = {Song, Jifei and Song, Yi-Zhe and Xiang, Tony and Hospedales, Timothy},
  booktitle={Proceedings of the British Machine Vision Conference (BMVC)},
  pages={45.1-45.12},
  year = {2017}
}

@inproceedings{sangkloy2022sketch,
  title={A Sketch Is Worth a Thousand Words: Image Retrieval with Text and Sketch}, 
  author={Patsorn Sangkloy and Wittawat Jitkrittum and Diyi Yang and James Hays},
  booktitle={European Conference on Computer Vision (ECCV)},
  pages={251-267},
  year={2022}
}

@inproceedings{hu2022feature,
  title={Feature representation learning for unsupervised cross-domain image retrieval},
  author={Hu, Chonghui and Lee, Gim Hee},
  booktitle={European Conference on Computer Vision (ECCV)},
  year={2022}
}

@article{wang2023correspondence,
  title={Correspondence-Free Domain Alignment for Unsupervised Cross-Domain Image Retrieval}, 
  volume={37}, 
  DOI={10.1609/aaai.v37i8.26215}, 
  number={8}, 
  journal={Proceedings of the AAAI Conference on Artificial Intelligence}, 
  author={Wang, Xu and Peng, Dezhong and Yan, Ming and Hu, Peng}, 
  year={2023}, 
  pages={10200-10208} 
}

@inproceedings{li2021prototypical,
    title={Prototypical Contrastive Learning of Unsupervised Representations},
    author={Junnan Li and Pan Zhou and Caiming Xiong and Steven Hoi},
    booktitle={International Conference on Learning Representations (ICLR)},
    year={2021},
}

@inproceedings{vendrow2024inquire,
 author = {Vendrow, Edward and others},
 booktitle = {Advances in Neural Information Processing Systems},
 doi = {10.52202/079017-4018},
 editor = {A. Globerson and L. Mackey and D. Belgrave and A. Fan and U. Paquet and J. Tomczak and C. Zhang},
 pages = {126500--126514},
 publisher = {Curran Associates, Inc.},
 title = {INQUIRE: A Natural World Text-to-Image Retrieval Benchmark},
 volume = {37},
 year = {2024}
}



\end{document}